\documentclass[11pt,a4paper]{article}
\usepackage[hyperref]{emnlp2018}
\usepackage{times}
\usepackage{latexsym}
\usepackage{stmaryrd}
\usepackage{graphicx}
\usepackage{amsmath}
\usepackage{array,multirow}
\usepackage{comment}
\usepackage{enumitem}
\usepackage[ruled]{algorithm2e}
\SetAlFnt{\small}
\SetAlCapFnt{\small}
\SetAlCapNameFnt{\small}

\newcolumntype{?}{!{\vrule width 1pt}}
\newcolumntype{@}{!{\vrule width 1.5pt}}

\usepackage{url}
\aclfinalcopy

\title{Learning to Learn Semantic Parsers from\\ Natural Language Supervision}

\author{Igor Labutov\thanks{\;\;Work done while at Carnegie Mellon University.}\\
  LAER AI, Inc.\\
  New York, NY\\
  {\tt igor.labutov@laer.ai} \\\And
  Bishan Yang\footnotemark[1]\\
  LAER AI, Inc.\\
  New York, NY\\
  {\tt bishan.yang@laer.ai} \\
  \\\And
  Tom Mitchell\\
  Carnegie Mellon University\\
  Pittsburgh, PA\\
  {\tt tom.mitchell@cmu.edu} \\}

\date{}

\begin{document}
\maketitle
\begin{abstract}
As humans, we often rely on language to learn language. For example, when corrected in a conversation, we may learn from that correction, over time improving our language fluency. Inspired by this observation, we propose a learning algorithm for training semantic parsers from supervision (feedback) expressed in natural language. Our algorithm learns a semantic parser from users' corrections such as \textit{``no, what I really meant was before his job, not after''}, by also simultaneously learning to parse this natural language feedback in order to leverage it as a form of supervision. Unlike supervision with gold-standard logical forms, our method does not require the user to be familiar with the underlying logical formalism, and unlike supervision from denotation, it does not require the user to know the correct answer to their query. This makes our learning algorithm naturally scalable in settings where existing conversational logs are available and can be leveraged as training data. We construct a novel dataset of natural language feedback in a conversational setting, and show that our method is effective at learning a semantic parser from such natural language supervision.
\end{abstract}

\section{Introduction}
\label{sec:introduction}
Semantic parsing is a problem of mapping a natural language utterance into a formal meaning representation, e.g., an executable logical form~\cite{zelle1996learning}. Because the space of all logical forms is large but constrained by an underlying structure (i.e., all trees), the problem of learning a semantic parser is commonly formulated as an instance of \textit{structured prediction}. 

Historically, approaches based on \textit{supervised learning} of structured prediction models have emerged as some of the first and still remain common in the semantic parsing community ~\cite{zettlemoyer2005learning,zettlemoyer2009learning,kwiatkowski2010inducing}. A well recognized practical challenge in supervised learning of structured models is that fully annotated structures (e.g., logical forms) that are needed for training are often highly labor-intensive to collect. This problem is further exacerbated in semantic parsing by the fact that these annotations can only be done by people familiar with the underlying logical language, making it challenging to construct large scale datasets by non-experts. 

Over the years, this practical observation has spurred many creative solutions to training semantic parsers that are capable of leveraging weaker forms of supervision, amenable to non-experts. One such weaker form of supervision relies on logical form denotations (i.e, the results of a logical form's execution) -- rather than the logical form itself, as ``supervisory'' signal~\cite{clarke2010driving,liang2013learning,berant2013semantic,pasupat2015compositional,liang2016neural,krishnamurthy2017neural}. In Question Answeing (QA), for example, this means the annotator needs only to know the answer to a question, rather than the full SQL query needed to obtain that answer. Paraphrasing of utterances already annotated with logical forms is another practical approach to scale up annotation without requiring experts with a knowledge of the underlying logical formalism~\cite{liangpara1,liangpara2}.

\begin{figure}
\includegraphics[width=7.7cm]{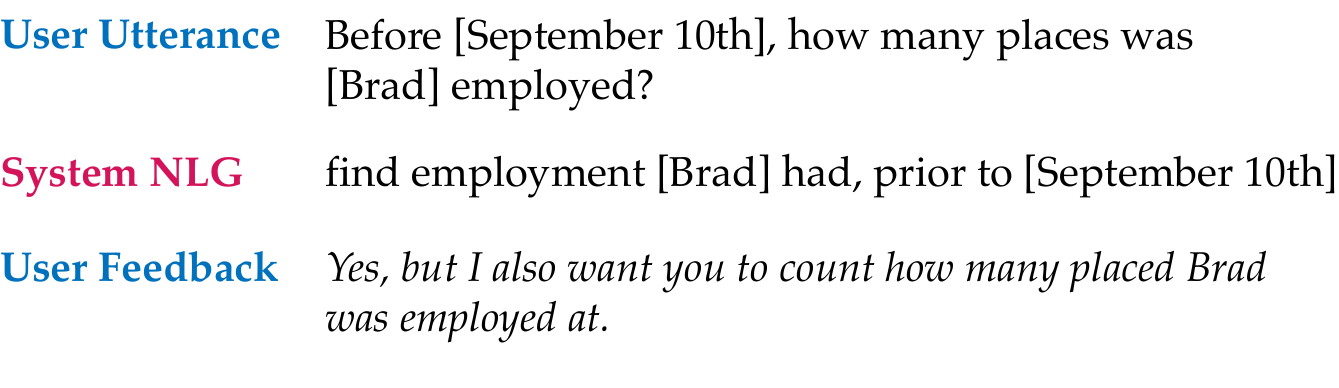}
\centering
\caption{Example (i) user's original utterance, (ii) confirmation query generated by inverting the original parse, (iii) user's generated feedback towards the confirmation query (i.e., original parse) .}
\label{fig:examples}
\end{figure}

Although these and similar methods do reduce the difficulty of the annotation task, collecting even these weaker forms of supervision (e.g., denotations and paraphrases) still requires a dedicated annotation event, that occurs outside of the normal interactions between the end-user and the semantic parser\footnote{although in some cases existing datasets can be leveraged to extract this information}. Furthermore, expert knowledge may still be required, even if the underlying logical form does not need to be given by the annotator (QA denotations, for example, require the annotator to know the correct answer to the question -- an assumption which doesn't hold for end-users who asked the question with the goal of obtaining the answer). In contrast, our goal is to leverage natural language feedback and corrections that may occur naturally as part of the continuous interaction with the non-expert end-user, as training signal to learn a semantic parser. 


The core challenge in leveraging natural language feedback as a form of supervision in training semantic parsers, however, is the challenge of correctly parsing that feedback to extract the supervisory signal embedded in it. Parsing the feedback, just like parsing the original utterance, requires its own semantic parser trained to interpret that feedback. Motivated by this observation, our main contribution in this work is a semi-supervised learning algorithm that learns a task parser (e.g., a question parser) from feedback utterances while simultaneously learning a parser to interpret the feedback utterances. Our algorithm relies only on a small number of annotated logical forms, and can continue learning as more feedback is collected from interactions with the user.  

Because our model learns from supervision that it simultaneously learns to interpret, we call our approach \textit{learning to learn semantic parsers from natural language supervision}.
%
%

\section{Problem Formulation}
Formally, the setting proposed in this work can be modelled as follows: (i) the user poses a natural language input $u_i$ (e.g.,  a question) to the system, (ii) the system parses the user's utterance $u_i$, producing a logical form $\hat{y}_i$, (iii) the system communicates $\hat{y}_i$ to the user in natural language in the form of a confirmation (i.e., \textit{``did you mean \ldots''}), (iv) in response to the system's confirmation, the user generates a feedback utterance $f_i$, which may be a correction of $\hat{y}_i$ expressed in natural language. The observed variables in a single interaction are the task utterance $u_i$, the predicted task logical form $\hat{y}_i$ and the user's feedback $f_i$;  the true logical form $y_i$ is hidden. See Figure \ref{fig:examples} for an illustration.

A key observation that we make from the above formulation is that learning from such interactions can be effectively done in an \textit{offline} (i.e., non-interactive) setting, using only the logs of past interactions with the user. Our aim is to formulate a model that can learn a task semantic parser (i.e., one parsing the original utterance $u$) from such interaction logs, without access to the true logical forms (or denotations) of the users' requests.

\subsection{Modelling conversational logs}
\label{sec:conv-logs}

Formally, we propose to learn a semantic parser from conversational logs represented as follows:
\vspace{-1mm}
$$
D = \{ (u_i,\hat{y}_i,f_i) \}_{i, \ldots, N}
$$
where $u_i$ is the user's task utterance (e.g., a question), $\hat{y}_i$ is the system's original parse (logical form) of that utterance, $f_i$ is the user's natural language feedback towards that original parse and $N$ is the number of dialog turns in the log. Note that the original parse $\hat{y}_i$ of utterance $u_i$ could come from any semantic parser that was deployed at the time the data was logged -- there are no assumptions made on the source of how $\hat{y}_i$ was produced.

Contrast this with the traditional learning settings for semantic parsers, where the user (or annotator) provides the correct logical form $y_i$ or the execution of the correct logical form $\llbracket y_i \rrbracket$ (denotation) (Table~\ref{table:supervision}). In our setting, instead of the correct logical form $y_i$, we only have access to the logical form produced by whatever semantic parser was interacting with the user at the time the data was collected, i.e.,  $\hat{y}_i$ is \textit{not} necessarily the correct logical form (though it could be). In this work, however, we focus only on corrective feedback (i.e., cases where $\hat{y} \neq y$) as our main objective is to evaluate the ability to interpret rich natural language feedback (which is only given when the original parse is incorrect). Our key hypothesis in this work is that although we do not observe $y_i$ directly, combining $\hat{y}_i$ with the user's feedback $f_i$ should be sufficient to get a better estimate of the correct logical form (closer to $y_i$), which we can then leverage as a ``training example'' to improve the task parser (and the feedback parser). 

\begin{table}
\centering
\small
\begin{tabular}{| l |l |}
\hline
  \textbf{Supervision} & \textbf{Dataset}  \\ \hline
  Full logical forms & $\{ (u_i,y_i) \}_{i, \ldots, N}$ \\ 
  Denotations & $\{ (u_i, \llbracket y_i \rrbracket) \}_{i, \ldots, N}$\\ 
  Binary feedback & $\{ (u_i, \llbracket \hat{y}_i = y_i \rrbracket) \}_{i, \ldots, N}$ \\ 
  \hline
  NL feedback (this work) & $\{ (u_i,\hat{y}_i,f_i) \}_{i, \ldots, N}$ \\  \hline
\end{tabular}
\caption{Different types of supervision used in literature for training semantic parsers, and the corresponding data needed for each type of supervision. Notation used in the table: $u$ corresponds to user utterance (language), $y$ corresponds to a gold-standard logical form parse of $u$, $\hat{y}$ corresponds to a predicted logical form, $\llbracket \cdot \rrbracket$ is the result of executing an expression inside the brackets and $f$ is the user's feedback expressed in natural language in the context of an utterance and a predicted logical form.}
\label{table:supervision}
\end{table}

\section{Natural Language Supervision}
\label{sec:learning}

\subsection{Learning problem formulation}
In this section, we propose a learning algorithm for training a semantic parser from natural language feedback. We will use the terms \textit{task parser} and \textit{feedback parser} to refer to the two distinct parsers used for parsing the user's original task utterance (e.g., question) and parsing the user's follow-up feedback utterance respectively. Our learning algorithm does not assume any specific underlying model for the two parsers aside from the requirement that each parser specifies a probability distribution over logical forms given the utterance (for the task parser) and the utterance plus feedback (for the feedback parser): 

\vspace{2mm}

\textbf{Task Parser}: $P(y \mid u; \theta_t)$

\textbf{Feedback Parser}: $P(y \mid u, f, \hat{y} ; \theta_f)$

\vspace{2mm}

\noindent where $\theta_t$ and $\theta_f$ parametrize the task and feedback parsers respectively. Note that the key distinction between the task and feedback parser models is that in addition to the user's original task utterance $u$, the feedback parser also has access to the user's feedback utterance $f$, and the original parser's prediction $\hat{y}$. We now introduce a joint model that combines task and feedback parsers in a way that encourages the two models to agree:
\begin{equation*}
\small
\begin{split}
P(y \mid u, f, \hat{y}; \theta_t, \theta_f) = \frac{1}{Z}  \underbrace{P(y \mid u; \theta_t)}_{\textup{task parser}} \underbrace{P(y \mid u, f, \hat{y} ; \theta_f)}_{\textup{feedback parser}}
\end{split}
\end{equation*}
\noindent At training time, our objective is to maximize the above joint likelihood by optimizing the parser parameters $\theta_t$ and $\theta_f$ of the task and feedback parsers respectively. The intuition behind optimizing the joint objective is that it encourages the ``weaker'' model that does not have access to the feedback utterance to agree with the ``stronger'' model that does (i.e., using the feedback parser's prediction as a noisy ``label'' to bootstrap the task parser), while conversely encouraging a more ``complex'' model to agree with a ``simpler'' model (i.e., using the simpler task parser model as a ``regularizer'' for the more complex feedback parser; note that the feedback parser generally has higher model complexity compared to the task parser because it incorporates additional parameters to account for processing the feedback input).\footnote{Note that in practice, to avoid locally optimal degenerate solutions (e.g., where the feedback parser learns to trivially agree with the task parser by learning to ignore the feedback), some amount of labeled logical forms would be required to pre-train both models.} Note the feedback parser output is not simply the meaning of the feedback utterance in isolation, but instead the revised semantic interpretation of the original utterance, guided by the feedback utterance.  In this sense, the task faced by the feedback parser is to both determine the meaning of the feedback, and to apply that feedback to repair the original interpretation of $u_i$. Note that this model is closely related to co-training \cite{cotraining} (and more generally multi-view learning \cite{multiview}) and is also a special case of a product-of-experts (PoE) model~\cite{poe}.

\subsection{Learning}
The problem of maximizing the joint-likelihood $P(y \mid u, f, \hat{y}; \theta_t, \theta_f)$ can be approached as a standard problem of maximum likelihood estimation in the presence of latent variables (i.e., unobserved logical form $y$), suggesting the application of the Expectation Maximization (EM) algorithm for learning parameters $\theta_t$ and $\theta_f$. The direct application of EM in our setting, however, is faced with a complication in the E-step. Because the hidden variables (logical forms $y$) are structured, computing the posterior over the space of all logical forms and taking the expectation of the log-likelihood with respect to that posterior is generally intractable (unless we assume certain factorized forms for the logical form likelihood).

Instead, we propose to approximate the E-step by replacing the expectation of the log joint-likelihood with its point estimate, using the maximum a posteriori (MAP) estimate of the posterior over logical forms $y$ to obtain that estimate (this is sometimes referred to as ``hard-EM''). Obtaining the MAP estimate of the posterior over logical forms may itself be a difficult optimization problem, and the specific algorithm for obtaining it would depend on the internal models of the task and the feedback parser. 

As an alternative, we propose a simple, MCMC based algorithm for approximating the MAP estimate of the posterior over logical forms, that does not require access to the internals of the parser models, i.e., allowing us to conveniently treat both parsers as ``black boxes''. The only assumption that we make is that it's easy to sample logical forms from the individual parsers (though not necessarily from the joint model), as well as to compute the likelihoods of those sampled logical forms under at least one of the models. 

We use Metropolis Hastings to sample from the posterior over logical forms, using one of the parser models as the proposal distribution. Specifically, if we choose the feedback parser as the proposal distribution, and initialize the Markov Chain with a logical form sampled from the task parser, it can be shown that the acceptance probability $r$ for the first sample conveniently simplifies to the following expression:
\begin{equation}
\small
    r = \min \left (1, \frac{P(\hat{y}_f \mid u, \theta_t)}{P(\hat{y}_{t} \mid u, \theta_t)} \right )
\end{equation}
where $\hat{y}_t$ and $\hat{y}_f$ are logical forms sampled from the task and feedback parsers respectively. Intuitively the above expression compares the likelihood of the logical form proposed by the task parser to the likelihood of the logical form proposed by the feedback parser, but with \textit{both} likelihoods computed under the same \textit{task} parser model (making the comparison fair). The proposed parse is then accepted with a probability proportional to that ratio. See Algorithm~\ref{alg:alg2} for details.

Finally, given this MAP estimate of $y$, we can perform optimization over the parser parameters $\theta_t$ and $\theta_f$ using a single step with stochastic gradient descent before re-estimating the latent logical form $y$. We also approximate the gradients of each parser model by ignoring the gradient terms associated with the log of the normalizer $Z$. \footnote{We have experimented with a sampling based approximation of the true gradients with contrastive divergence \cite{contrastive},  however, found that our approximation works sufficiently well empirically. See Algorithm~\ref{alg:alg1} for more details of the complete algorithm. }

\begin{algorithm}[ht!]
\label{alg:alg1}
\SetKwFunction{Sample}{Sample}
\SetKwFunction{MH}{MH-MAP}
\SetKwFunction{SGD}{SGD}
\SetKwInOut{Input}{Input}\SetKwInOut{Output}{Output}
\SetKwInOut{Parameter}{Parameter}

 \Input{$D = \{(u_i, \hat{y}_i, f_i)\}_{1, \ldots, N}$}
 \Output{Task parser parameters $\theta_t$; Feedback parser parameters $\theta_f$}
 \Parameter{Number of training epochs $T$}
 \For{$t = 1$ \KwTo $T$}{
 \For{$i = 1$ \KwTo $N$}{
   $\hat{y}^f_i \gets$ \MH($u_i$, $\hat{y}_i$, $f_i$, $\theta_t$, $\theta_f$) \;
   $\nabla \theta_t \gets \nabla  \log P(\hat{y}^f_i \mid u_i; \theta_t)$ \;
   $\nabla \theta_f \gets \nabla  \log P(\hat{y}^f_i \mid u_i, f_i, \hat{y}_i; \theta_f)$ \;
   $\theta_t \gets$ \SGD($\theta_t, \nabla \theta_t$) \;
   $\theta_f \gets$ \SGD($\theta_f, \nabla \theta_f$) \;
   }
  }
 \caption{Semantic Parser Training from Natural Language Supervision}
\end{algorithm}

\begin{algorithm}[ht!]
\label{alg:alg2}
\SetKwInOut{Input}{Input}\SetKwInOut{Output}{Output}
\SetKwInOut{Parameter}{Parameter}

 \Input{$u$, $\hat{y}$, $f$, $\theta_t$, $\theta_f$}
 \Output{latent parse $\hat{y}^f$}
 \Parameter{Number of sampling iterations $N$}
\SetKwFunction{Sample}{Sample}
\SetKwFunction{AcceptRatio}{AcceptRatio}
\DontPrintSemicolon
  \SetKwFunction{FMain}{MH-MAP}
  \SetKwProg{Fn}{Function}{:}{}
    \Fn{\FMain{$u$, $\hat{y}$, $f$, $\theta_t$, $\theta_f$}}{
         $samples \gets [~]$ \;
        \emph{// sample parse from task parser} \;
        \Sample~$\hat{y}_t \sim P(y \mid u, \theta_t)$ \;
        $\hat{y}_{curr} \gets \hat{y}_t$ \;
        \For{$i = 1$ \KwTo $N$}{
            \emph{// sample parse from feedback parser} \;
            \Sample~$\hat{y}_f \sim P(y \mid u, f, \hat{y}, \theta_f)$ \;
            $r \gets \textup{min} \left (1, \frac{P(\hat{y}_f \mid u, \theta_t)}{P(\hat{y}_{curr} \mid u, \theta_t)} \right )$ \;
            \Sample~$accept \sim \textup{Bernoulli}(r)$ \;
             \If{accept}{
                $p_t \gets P(\hat{y}_f \mid u, \theta_t)$ \;
                $p_f \gets P(\hat{y}_f \mid u, f, \hat{y}, \theta_f)$ \;
                $samples[\hat{y}_{f}] \gets p_t \cdot p_f$ \;
                $\hat{y}_{curr} \leftarrow \hat{y}_f$ \;
             }
        }
        $\hat{y}^f \gets \textup{argmax} ~samples$ \;
        \KwRet $\hat{y}^f$ \;
  }
  
 \caption{Metropolis Hastings-based MAP estimation of latent semantic parse}
\end{algorithm}

\subsection{Task Parser Model}
Our task parser model is implemented based on existing attention-based encoder-decoder models~\cite{bahdanau2014neural,luong2015effective}. The encoder takes in an utterance (tokenized) $u$ and computes a context-sensitive embedding $h_i$ for each token $u_i$ using a bidirectional Long Short-Term Memory (LSTM) network~\cite{hochreiter1997long}, where $h_i$ is computed as the concatenation of the hidden states at position $i$ output by the forward LSTM and the backward LSTM. The decoder generates the output logical form $y$ one token at a time using another LSTM. At each time step $j$, it generates $y_j$ based on the current LSTM hidden state $s_j$, a summary of the input context $c_j$, and attention scores $a_{ji}$, which are used for attention-based copying as in~\cite{jia2016data}. Specifically, $$p(y_j=w, w\in V_{out}|u, y_{1:j-1})\propto \exp\left(W_o[s_j;c_j]\right)$$
$$p(y_j=u_i|u, y_{1:j-1})\propto \exp\left(a_{ji}\right)$$
where $y_j=w$ denotes that $y_j$ is chosen from the output vocabulary $V_{out}$; $y_j=u_i$ denotes that $y_j$ is a copy of $u_i$; $a_{ji}=s_j^TW_ah_i$ is an attention score on the input word $u_i$; $c_j=\sum_i\alpha_i h_i, \alpha_i\propto\exp\left(a_{ji}\right)$ is a context vector that summarizes the encoder states; and $W_o$ and $W_a$ are matrix parameters to be learned. After generating $y_j$, the decoder LSTM updates its hidden state $s_{j+1}$ by taking as input the concatenation of the embedding vector for $y_{j}$ and the context vector $c_j$. 

An important problem in semantic parsing for conversation is resolving references of people and things mentioned in the dialog context. Instead of treating coreference resolution as a separate problem, we propose a simple way to resolve it as a part of semantic parsing. For each input utterance, we record a list of previously-occurring entity mentions $m=\{m_1,...,m_L\}$. We consider entity mentions of four types: persons, organizations, times, and topics. The encoder now takes in an utterance $u$ concatenated with $m$\footnote{The mentions are ordered based on their types and each mention is wrapped by special boundary symbols $[$ and $]$.}, and the decoder can generate a referenced mention through copying. 

The top parts of Figure~\ref{fig:combined_1} and Figure~\ref{fig:combined_4} visualize the attention mechanism of the task parser, where the decoder attends to both the utterance and the conversational context during decoding. Note that the conversational context is only useful when the utterance contains reference mentions.

\subsection{Feedback Parser Model}
Our feedback parser model is an extension of the encoder-decoder model in the previous section. The encoder consists of two bidirectional LSTMs: one encodes the input utterance $u$ along with the history of entity mentions $m$ and the other encodes the user feedback $f$. At each time step $j$ during decoding, the decoder computes attention scores over each word $u_i$ in utterance $u$ as well as each feedback word $f_k$ based on the decoder hidden state $s_j$, the context embedding $b_k$ output by the feedback encoder, and a learnable weight matrix $W_e$: $e_{jk}=s_j^TW_eb_k$. The input context vector $c_j$ in the previous section is updated by adding a context vector that attends to both the utterance and the feedback: 
$$c'_j=\sum_i\alpha_i h_i+\sum_k \beta_k b_k$$
where $\alpha_i\propto \exp(a_{ji})$ and $\beta_k\propto \exp(e_{jk})$. Accordingly, the decoder is allowed to copy words from the feedback:
$$p(y_j=f_k|u,f,y_{1:j-1})\propto \exp\left(e_{jk}\right)$$

The bottom parts of Figure~\ref{fig:combined_1} and Figure~\ref{fig:combined_4} visualize the attention mechanism of the feedback parser, where the decoder attends to the utterance, the conversational context, and the feedback during decoding.

Note that our feedback model does not explicitly incorporate the original logical form $\hat{y}$ that the user was generating their feedback towards. We experimented with a number of ways to incorporate $\hat{y}$ in the feedback parser model, but found it most effective to instead use it during MAP inference for the latent parse $y$. In sampling a logical form in Algorithm~\ref{alg:alg2}, we simply reject the sample if the sampled logical form matches $\hat{y}$.

\section{Dataset}
\label{sec:dataset}
In this work we focus on the problem of semantic parsing in a conversational setting and construct a new dataset for this task. Note that a parser is required to resolve references to the conversational context during parsing, which in turn may further amplify ambiguity and propagate errors to the final logical form. Conveniently, the same conversational setting also offers a natural channel for correcting such errors via natural language feedback users can express as part of the conversation. 


\subsection{Dataset construction}

We choose \textit{conversational search} in the domain of email and biographical research as a setting for our dataset. Large enterprises produce large amounts of text during daily operations, e.g., emails, reports, and meeting memos. There is an increasing need for systems that allow users to quickly find information over text through search. Unlike regular tasks like booking airline tickets, search tasks often involve many facets, some of which may or may not be known to the users when the search begins. For example, knowing where a person works may triggers a followup question about whether the person communicates with someone internally about certain topic. Such search tasks will often be handled most naturally through dialog.

\begin{figure}
\includegraphics[width=6.5cm]{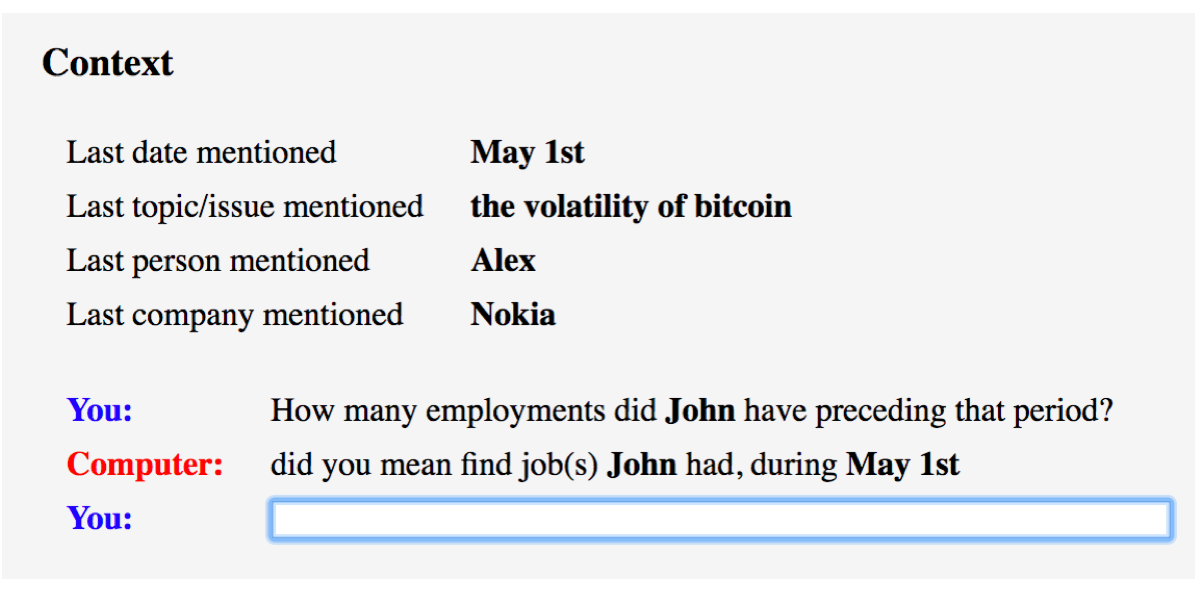}
\centering
\caption{Screenshot of the Mechanical Turk web interface used to collect natural language feedback.}
\label{fig:screenshot}
\end{figure}

Figure~\ref{fig:examples} shows example dialog turns containing the user's utterance (question) $u$, system's confirmation of the original parse $\hat{y}$, and the user's feedback $f$. To simplify and scale data collection, we decouple the dialog turns and collect feedback for each dialog turn in isolation. We do that by showing a worker on Mechanial Turk the original utterance $u$, the system's natural language confirmation generated by inverting the original logical form $\hat{y}$, and the dialog context summarized in a table. See Figure~\ref{fig:screenshot} for a screenshot of the Mechanical Turk interface. The dialog context is summarized by the entities (of types person, organization, time and topic) that were mentioned earlier in the conversation and could be referenced in the original question $u$ shown on the screen\footnote{zero or more of the context entities may actually be referenced in the original utterance}. The worker is then asked to type their feedback given $(u, \hat{y})$ and the dialog context displayed on the screen. Turkers are instructed to write a natural language feedback utterance that they might otherwise say in a real conversation when attempting to correct another person's incorrect understanding of their question. 

We recognize that our data collection process results in only an approximation of a true multi-turn conversation, however we find that this approach to data collection offers a convenient trade-off for collecting a large number of controlled and diverse context-grounded interactions. Qualitatively we find that turkers are generally able to imagine themselves in the hypothetical ongoing dialog, and are able to generate realistic contextual feedback utterances using only the context summary table provided to them.

The initial dataset of questions paired with original logical form parses $\{u_i, \hat{y}_i\}_{1, \ldots, N}$ that we use to solicit feedback from turkers, is prepared offline. In this separate offline task we collect a dataset of 3556 natural language questions, annotated with gold standard logical forms, in the same domain of email and biographical research. We parse each utterance in this dataset with a floating grammar-based semantic parser trained using a structured perceptron algorithm (implemented in SEMPRE~\cite{berant2013semantic}) on a subset of the questions. We then construct the dataset for feedback collection by sampling the logical form $\hat{y}$ from the first three candidate parses in the beam produced by the grammar-based parser. We use this grammar-based parser intentionally as a very different model from the one that we would ultimately train (LSTM-based parser) on the conversational logs produced by the original parser.

We retain 1285 out of the 3556 annotated questions to form a test set. The rest 2271 questions we pair with between one and three predicted parses $\hat{y}$ sampled from the beam produced by the grammar-based parser, and present each pair of original utterance and predicted logical form $(u_i,\hat{y}_i)$ to a turker who then generates a feedback utterance $f_i$. In total, we collect 4321 question/original parse/feedback triples $(u_i, \hat{y}_i, f_i)$ (averaging approximately 1.9 feedback utterances per question).

\section{Experiments}
The key hypothesis that we aim to evaluate in our experiments is whether natural language feedback is an effective form of supervision for training a semantic parser. To achieve this goal, we control and measure the effect that the number of feedback utterances used during training has on the resulting performance of the task semantic parser on a held-out test set. Across all our experiments we also use a small seed training set (300 questions) that contains gold-standard logical forms to pre-train the task and feedback parsers. The number of ``unlabeled" questions\footnote{Note that from hereon we will refer to the portion of the data that contains natural language feedback as the only form of supervision as ``unlabeled data'', to emphasize that it is not labeled with gold standard logical forms (in contrast to the seed training set).} (i.e., questions not labeled with gold standard logical forms but that have natural language feedback) ranges from 300, 500, 1000 to 1700 representing different experimental settings. For each experimental setting, we rerun the experiment 10 times, re-sampling the questions in both the training and unlabeled sets, and report the averaged results. The test-set remains fixed across all experiments and contains 1285 questions labeled with gold-standard logical forms. The implementation details for the task parser and the feedback parser are included in the appendix.
  


\subsection{Models and evaluation metrics}

In our evaluations, we compare the following four models:
\begin{itemize}[leftmargin=*]
    \item \textbf{MH (full model)} Joint model described in Section~\ref{sec:learning} and in Algorithm~\ref{alg:alg1} (using Metropolis Hastings-based MAP inference described in Algorithm~\ref{alg:alg2}).
    \item \textbf{MH (no feedback)} Same as the full model, except we ignore the feedback $f$ and the original logical form $\hat{y}$ information in the feedback model. Effectively, this reduces the model of the feedback parser to that of the task parser. Because during training, both models would be initialized differently, we may still expect the resulting model averaging effects to aid learning.
    \item \textbf{MH (no feedback + reject $\boldsymbol{\hat{y}}$)} Same as the above baseline without feedback, but we incorporate the knowledge of the original logical form $\hat{y}$ during training. We incorporate $\hat{y}$ using the same method as described in Section~\ref{sec:feedback-model}. 
    \item \textbf{Self-training} Latent logical form inference is performed using only the task parser (using beam search). Feedback utterance $f$ and original logical form $\hat{y}$ are ignored. Task parser parameters $\theta_t$ are updated in the same way as in Algorithm~\ref{alg:alg1}.
\end{itemize}
Note that all models are exposed to the same training seed set, and differ only in the way they take advantage of the unlabeled data. We perform two types of evaluations of each model:

\begin{itemize}[leftmargin=*]
    \item \textbf{Generalization performance} we use the learned task parser to make predictions on held-out data. This type of evaluation tests the ability of the parser trained with natural language feedback to generalize to unseen utterances.
    \item \textbf{Unlabeled data performance} we use the learned task parser to make predictions on the unlabeled data that was used in training it. Note that for each experimental condition, we perform this evaluation only on the portion of the unlabeled data that was used during training. This ensures that this evaluation tests the model's ability to ``recover'' correct logical forms from the questions that have natural language feedback associated with them. 
\end{itemize}

\section{Results}

\subsection{Generalization performance} 

Figure~\ref{fig:acc}\textit{a} shows test accuracy as a function of the number of unlabeled questions (i.e., questions containing only feedback supervision without gold standard logical forms) used during training, across all four models. As expected, using more unlabeled data generally improves generalization performance. The \textbf{self-training} baseline is the only exception, where performance starts to deteriorate as the ratio of unlabeled to labeled questions increases beyond a certain point. This behavior is not necessarily surprising -- when the unlabeled examples significantly outnumber the labeled examples, the model may more easily veer away to local optima without being strongly regularized by the loss on the small number of labeled examples.

Interestingly, the \textbf{MH (no feedback)} baseline is very similar to \textbf{self-training}, but has a significant performance advantage that does not deteriorate with more unlabeled examples. Recall that the \textbf{MH (no feedback)} model modifies the full model described in Algorithm~\ref{alg:alg1} by ignoring the feedback $f$ and the original logical form $\hat{y}$ in the feedback parser model $P(y \mid u, \hat{y}, f; \theta_f)$. This has the effect of reducing the model of the feedback parser into the model of the task parser $P(y \mid u; \theta_t)$. The training on unlabeled data proceeds otherwise in the same way as described in Algorithm~\ref{alg:alg1}. As a result, the principle behind the \textbf{MH (no feedback)} model is the same as that behind \textbf{self-training}, i.e., a single model learns from its own predictions. However, different initializations of the two copies of the task parser, and the combined model averaging appears to improve the robustness of the model sufficiently to keep it from diverging as the amount of unlabeled data is increased.

The \textbf{MH (no feedback + reject $\boldsymbol{\hat{y}}$)} baseline also does not observe the feedback utterance $f$, but incorporates the knowledge of the original parse $\hat{y}$. As described in Section~\ref{sec:feedback-model}, this knowledge is incorporated during MAP inference of the latent parse, by rejecting any logical form samples that match $\hat{y}$. As expected, incorporating the knowledge of $\hat{y}$ improves the performance of this baseline over the one that does not.

Finally, the full model that incorporates both, the feedback utterance $f$ and the original logical form $\hat{y}$ outperforms the baselines that incorporate only some of that information. The performance gain over these baselines grows as more questions with natural language feedback supervision are made available during training. Note that both this and the \textbf{MH (no feedback + reject $\boldsymbol{\hat{y}}$)} model incorporate the knowledge of the original logical form $\hat{y}$, however, the performance gain from incorporating the knowledge of $\hat{y}$ without the feedback is relatively small, indicating that the gains from the model that observes feedback is primarily from its ability to interpret it.

\subsection{Performance on unlabeled data}

\begin{figure}[ht!]
\includegraphics[width=6.0cm]{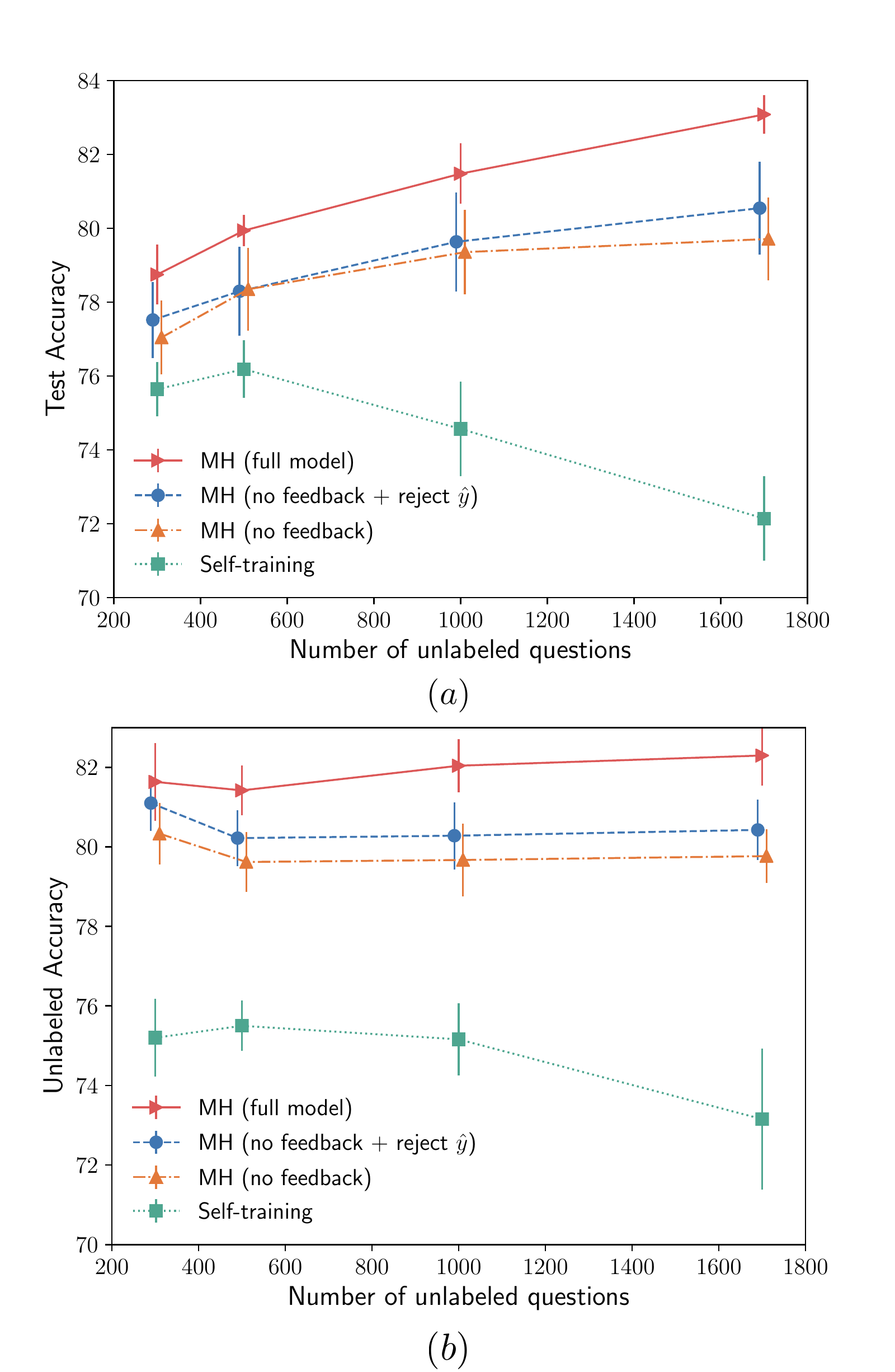}
\centering
\caption{\textit{(a)} Parsing accuracy on held-out questions as a function of the number of unlabeled questions used during training. \textit{(b)} Parsing accuracy on unlabeled questions as a function of the number of unlabeled questions used during training. In both panels, all parsers were initialized using 300 labeled examples consisting of questions and their corresponding logical form.} 
\label{fig:acc}
\end{figure}

Figure~\ref{fig:acc}\textit{b} shows accuracy on the unlabeled data, as a function of the number of unlabeled questions used during training, across all four models. The questions used in evaluating the model's accuracy on unlabeled data are the same unlabeled questions used during training in each experimental condition. The general trend and the relationship between baselines is consistent with the generalization performance on held-out data in Figure~\ref{fig:acc}\textit{a}. One of the main observations is that accuracy on unlabeled training examples remains relatively flat, but consistently high ($> 80\%$), across all models regardless of the amount of unlabeled questions used in training (within the range that we experimented with). This suggests that while the models are able to accurately recover the underlying logical forms of the unlabeled questions regardless of the amount of unlabeled data (within our experimental range), the resulting generalization performance of the learned models is significantly affected by the amount of unlabeled data (more is better).

\subsection{Effect of feedback complexity on performance}
%

\begin{figure}
\includegraphics[width=6.0cm]{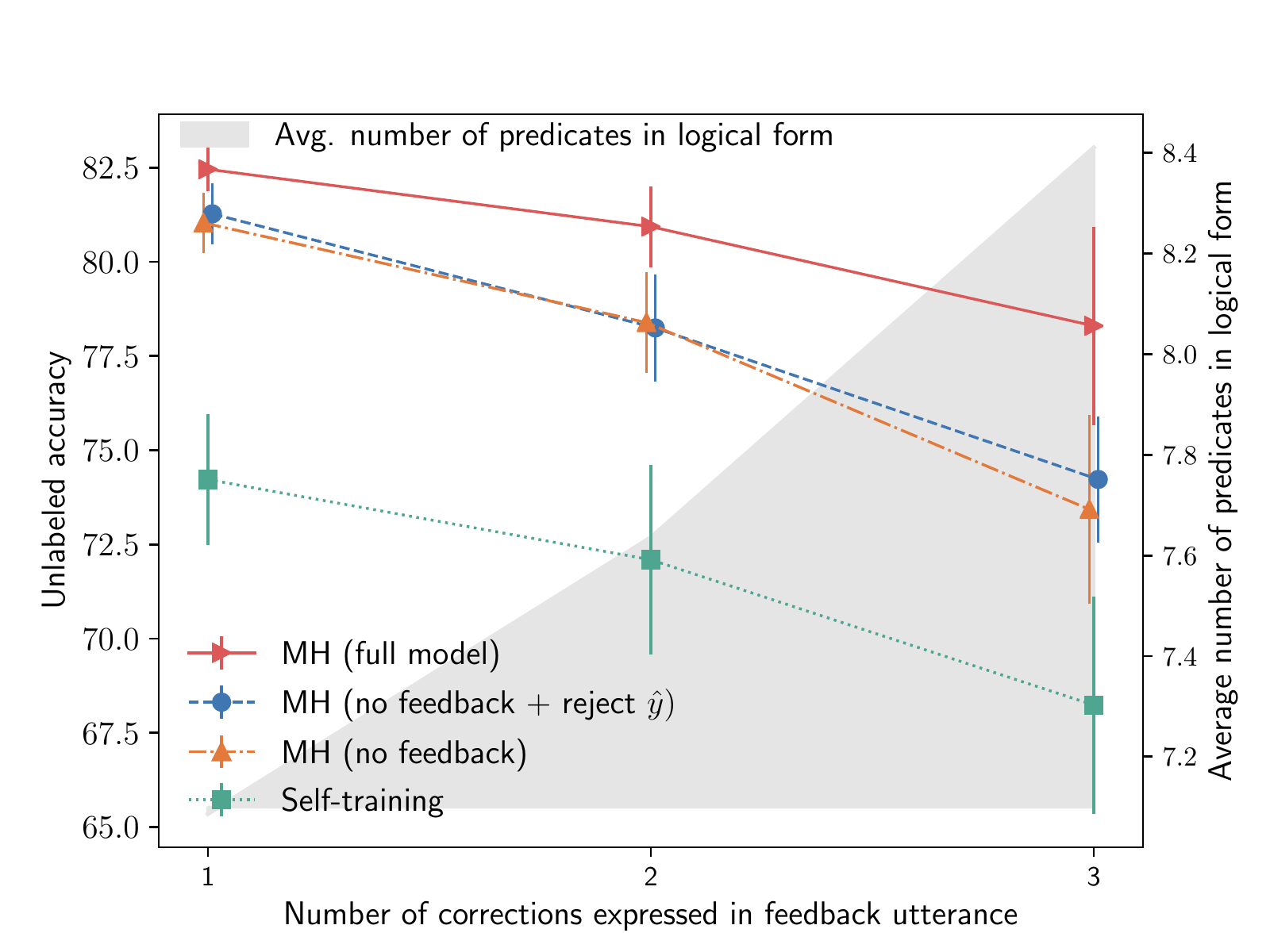}
\centering
\caption{Parsing accuracy on unlabeled questions, partitioned by feedback complexity (i.e., number of corrections expressed in a single feedback utterance).}
\label{fig:acc-breakdown}
\end{figure}

Figure~\ref{fig:acc-breakdown} reports parsing accuracy on unlabeled questions as a function of the number of corrections expressed in the feedback utterance paired with that question. Our main observation is that the performance of the full model (i.e., the joint model that uses natural language feedback) deteriorates for questions that are paired with more complex feedback (i.e., feedback containing more corrections). Perhaps surprisingly, however, is that all models (including those that do not incorporate feedback) deteriorate in performance for questions paired with more complex feedback. This is explained by the fact that more complex feedback is generally provided for more difficult-to-parse questions. In Figure~\ref{fig:acc-breakdown}, we overlay the parsing performance with the statistics on the average length of the target logical form (number of predicates). 

A more important take-away from the results in Figure~\ref{fig:acc-breakdown}, is that the model that takes advantage of natural language feedback gains an even greater advantage over models that do not use feedback when parsing more difficult questions. This means that the model is able to take advantage of more complex feedback (i.e., with more corrections) even for more difficult to parse questions.



%

\section{Related Work}
\label{sec:related}
Early semantic parsing systems map natural language to logical forms using inductive logical programming~\cite{zelle1996learning}. Modern systems apply statistical models to learn from pairs of sentences and logical forms~\cite{zettlemoyer2005learning,zettlemoyer2009learning,kwiatkowski2010inducing}. As hand-labeled logical forms are very costly to obtain, different forms of weak supervision have been explored. Example works include learning from pairs of sentences and answers by querying a database~\cite{clarke2010driving,liang2013learning,berant2013semantic,pasupat2015compositional,liang2016neural,krishnamurthy2017neural}; learning from indirect supervision from a large-scale knowledge base~\cite{reddy2014large,krishnamurthy2012weakly}; learning from conversations of systems asking for and confirming information~\cite{artzi2011bootstrapping,dialog-parser-1,dialog-parser-2}; and learning from interactions with a simulated world environment~\cite{branavan2009reinforcement,artzi2013weakly,goldwasser2014learning,misra2015environment}. The supervision used in these methods is mostly in the form of binary feedback, partial logical forms (e.g., slots) or execution results. In this paper, we explore a new form of supervision -- natural language feedback. We demonstrate that such feedback not only provides rich and expressive supervisory signals for learning but also can be easily collected via crowd-sourcing. Recent work~\cite{iyer2017learning} trains an online language-to-SQL parser from user feedback. Unlike our work, their collected feedback is structured and is used for acquiring more labeled data during training. Our model jointly learns from questions and feedback and can be trained with limited labeled data.


There has been a growing interest on machine learning from natural language instructions. Much work has been done in the setting where an autonomous agent learns to complete a task in an environment, for example, learning to play games by utilizing text manuals~\cite{branavan2012learning,eisenstein2009reading,narasimhan2015language} and guiding policy learning using high-level human advice~\cite{kuhlmann2004guiding,squiregrounding,harrison2017guiding}. Recently, natural language explanations have been used to augment labeled examples for concept learning~\cite{srivastava2017joint} and to help induce programs that solve algebraic word problems~\cite{ling2017program}. Our work is similar in that natural language is used as additional supervision during learning, however, our natural language annotations consist of user feedback on system predictions instead of explanations of the training data. 

\section{Conclusion and Future Work}

In this work, we proposed a novel task of learning a semantic parser directly from end-users' open-ended natural language feedback during a conversation. The key advantage of being able to learn from natural language feedback is that it opens the door to learning continuously through natural interactions with the user, but it also presents a challenge of how to interpret such feedback. 
In this work we introduced an effective approach that simultaneously learns two parsers: one parser that interprets natural language questions and a second parser that also interprets natural language feedback regarding errors made by the first parser.  

Our work is, however, limited to interpreting feedback contained in a single utterance. A natural generalization of learning from natural language feedback is to view it as part of an integrated dialog system capable of both interpreting feedback and asking appropriate questions to solicit this feedback (e.g., connecting to the work by \cite{dialog-parser-2}). We hope that the problem we introduce in this work, together with the dataset that we release, inspires the community to develop models that can learn language (e.g., semantic parsers) through flexible natural language conversation with end users.

 
\section*{Acknowledgements}

This research was supported in part by AFOSR under grant FA95501710218, and in part by the CMU InMind project funded by Oath.

\bibliography{naaclhlt2018}
\bibliographystyle{acl_natbib_nourl}
\newpage
\appendix

\section{Analysis of natural language feedback}

\begin{figure}
\includegraphics[width=7.5cm]{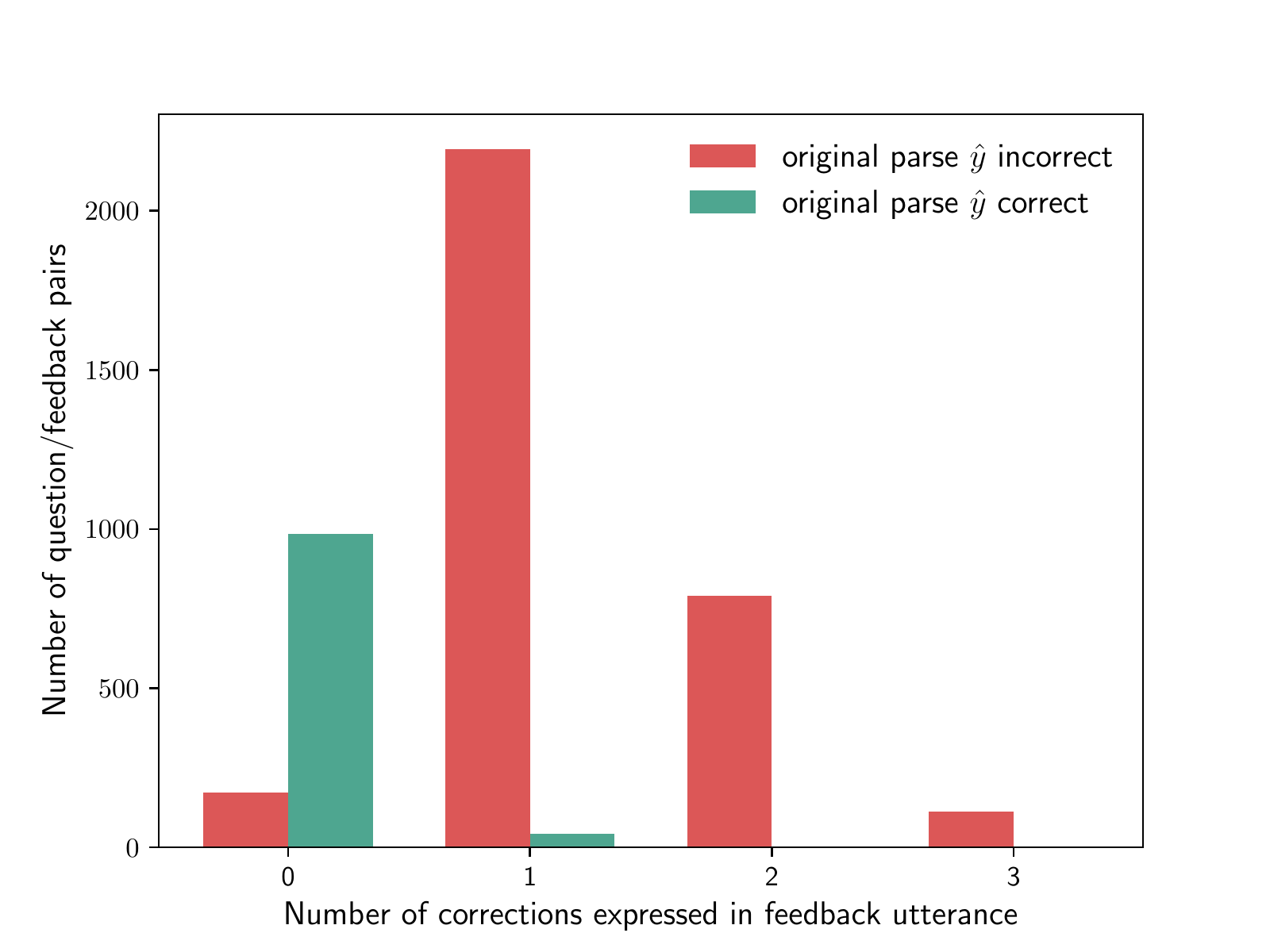}
\centering
\caption{A histogram of the number of corrections expressed in a natural language feedback utterance in our data (0 corrections means that the user affirmed the original parse as correct). We partition the feedback utterances according to whether the original parse $\hat{y}$ that the feedback was provided towards is known to be correct (i.e., matches the gold standard logical form parse).}
\label{fig:bar-plot}
\end{figure}

\begin{figure}[ht]
\includegraphics[width=7.5cm]{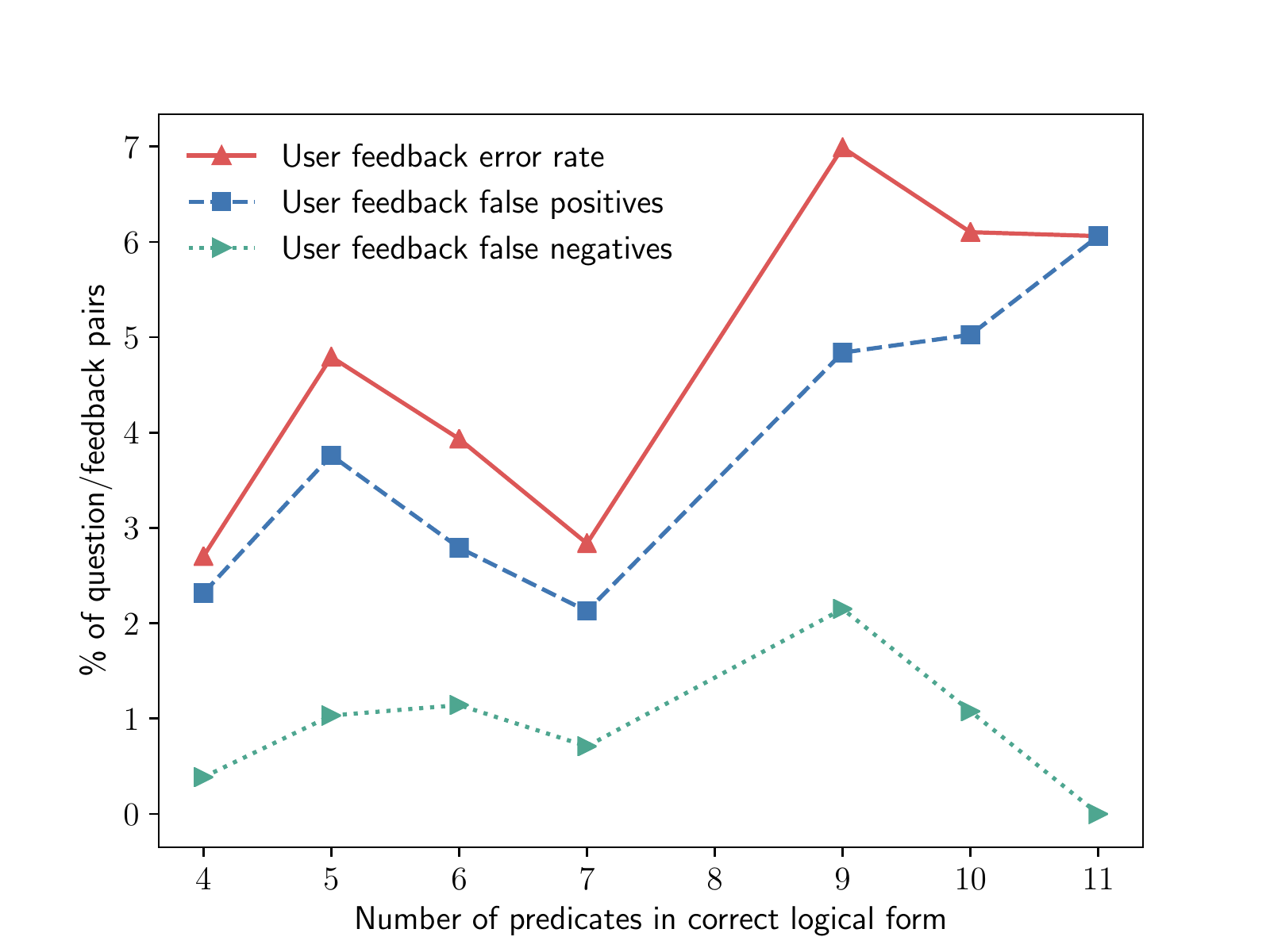}
\centering
\caption{Analysis of noise in feedback utterances contained in our dataset. \textit{Feedback false positives} refers to feedback that incorrectly identifies the wrong original parse $\hat{y}$ as correct. \textit{Feedback false negatives} refers to feedback that incorrectly identifies a correct original parse $\hat{y}$ as wrong. Users are more likely to generate false positives (i.e., miss the error) in their feedback for parses of more complex utterances (as measured by the number of predicates in the gold-standard logical form).}
\label{fig:feedback-noise}
\end{figure}

The dataset we collect creates a unique opportunity to study the nature and the limitations of the corrective feedback that users generate in response to an incorrect parse. One of our hypotheses stated in Section~\ref{sec:introduction} is that natural language affords users to express richer feedback than for example possible with binary (correct/incorrect) mechanism, by allowing users to explicitly refer to and fix what they see as incorrect with the original prediction. In this section we analyze the feedback utterances in our dataset to gain deeper insight into the types of feedback users generate, and the possible limitations of natural language as a source of supervision for semantic parsers.

Figure~\ref{fig:bar-plot} breaks down the collected feedback by the number of corrections expressed in the feedback utterance. The number of corrections ranges from $0$ (no corrections, i.e., worker considers the original parse $\hat{y}$ to be the correct parse of $u$) to more than 3 corrections.\footnote{Note that in cases where the user indicates that they made $0$ corrections, the feedback utterance that they write is often a variation of a confirmation such as \textit{``that's right''} or \textit{``that's correct''}} The number of corrections is self-annotated by the workers who write the feedback -- we instruct workers to count a correction constrained to a single predicate or a single entity as a single correction, and tally all such corrections in their feedback utterance after they have written it\footnote{we give these instructions to workers in an easy to understand explanation without invoking technical jargon}. Because we also know the ground truth of whether the original parse $\hat{y}$ (i.e., parse that the user provided feedback towards) was correct (i.e., $\hat{y}$ matches gold standard parse $y$), we can partition the number of corrections by whether the original parse $\hat{y}$ was correct (green bars in Figure~\ref{fig:bar-plot}) or incorrect (red bars), allowing us to evaluate the accuracy of some of the feedback. 

From Figure~\ref{fig:bar-plot}, we observe that users provide feedback that ranges in the number of corrections, with the majority of feedback utterances making one correction to the original logical form $\hat{y}$ (only 4 feedback utterances expressed more than 3 corrections). Although we do not have gold-standard annotation for the true number of errors in the original logical form $\hat{y}$, we can nevertheless obtain some estimate of the noise in natural language feedback by analyzing cases where we know that the original logical form $\hat{y}$ was correct, yet the user generated feedback with at least one correction. We will refer to such feedback instances as \textit{feedback false negatives}. Similarly, cases where the original logical form $\hat{y}$ is incorrect, yet the user provided no corrections in their feedback, we refer to as \textit{feedback false positives}. The number of feedback false negatives and false positives can be obtained directly from Figure~\ref{fig:bar-plot}. Generally, we observe that users are more likely to provide more false negatives ($\approx 4\%$ of all feedback) than false positives ($\approx 1\%$) in the feedback they generate.

It is also instructive to consider what factors may contribute to the observed noise in user generated natural language feedback. Our hypothesis is that more complex queries (i.e., longer utterances and longer logical forms) may result in a greater cognitive load to identify and correct the error(s) in the original parse $\hat{y}$. In Figure~\ref{fig:feedback-noise} we investigate this hypothesis by decomposing the percentage of feedback false positives and false negatives as a function of the number of predicates in the gold standard logical form (i.e., one that the user is trying to recover by making corrections in the original logical form $\hat{y}$). Our main observation is that users tend to generate more false positives (i.e., incorrectly identify an originally incorrect logical form $\hat{y}$ as correct) when the target logical form is longer (i.e., the query utterance $u$ is more complex). The number of false negatives (i.e., incorrectly identifying a correct logical form as incorrect) is relatively unaffected by the complexity of the query (i.e., number of predicates in the target logical form). One conclusion that we can draw from this analysis is that we can expect user-generated feedback to miss errors in more complex queries, and models that learn from users' natural language feedback need to have a degree of robustness to such noise. 

\section{Implementation Details}
We tokenize user utterances (questions) and feedback using the Stanford CoreNLP package. In all experiments, we use 300-dimensional word embeddings, initialized with word vectors trained using the Paraphrase Database PPDB~\cite{wieting2015towards}, and we use 128 hidden units for LSTMs. All parameters are initialized uniformly at random. We train all the models using Adam with initial learning rate $10^{-4}$ and apply L2 gradient norm clipping with a threshold of 10. In all the experiments, we pre-train the task parser and the feedback parser for 20 epochs, and then switch to semi-supervised training for 10 epochs. Pre-training takes roughly 30 minutes and the semi-supervised training process takes up to 7 hours on a Titan X GPU. 

%
\begin{figure*}
\includegraphics[width=0.8\paperwidth]{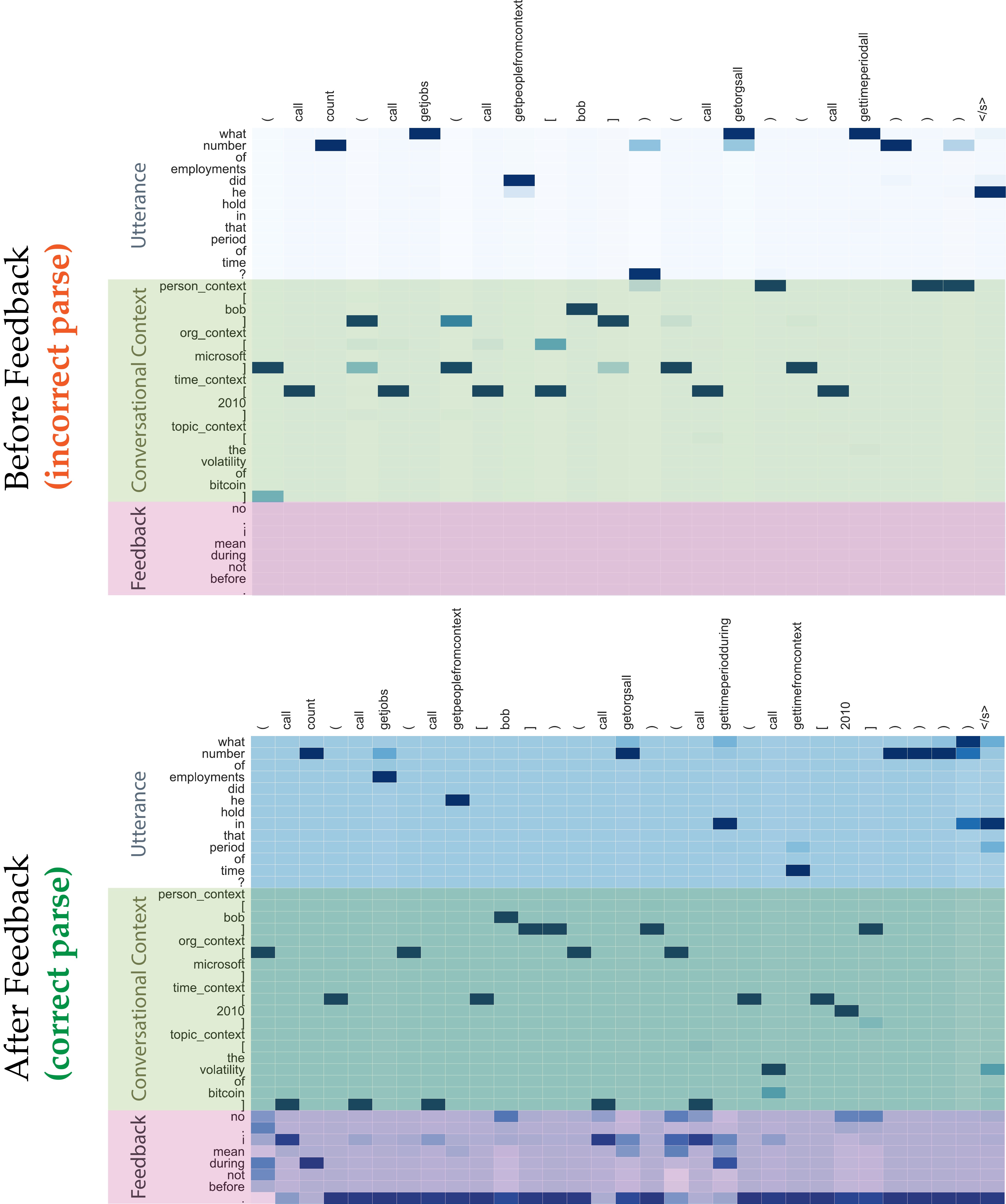}
\centering
\caption{Visualization of the attention mechanism in the task parser (top) and the feedback parser (bottom) for parsing the same utterance. We partition the input to the parser into three groups: the original utterance $u$ being parsed (blue), the conversational context (green) and the feedback utterance $f$ (red). This example was parsed incorrectly before incorporating feedback, but parsed correctly after its incorporation.}
\label{fig:combined_1}
\end{figure*}



\begin{figure*}
\includegraphics[width=0.8\paperwidth]{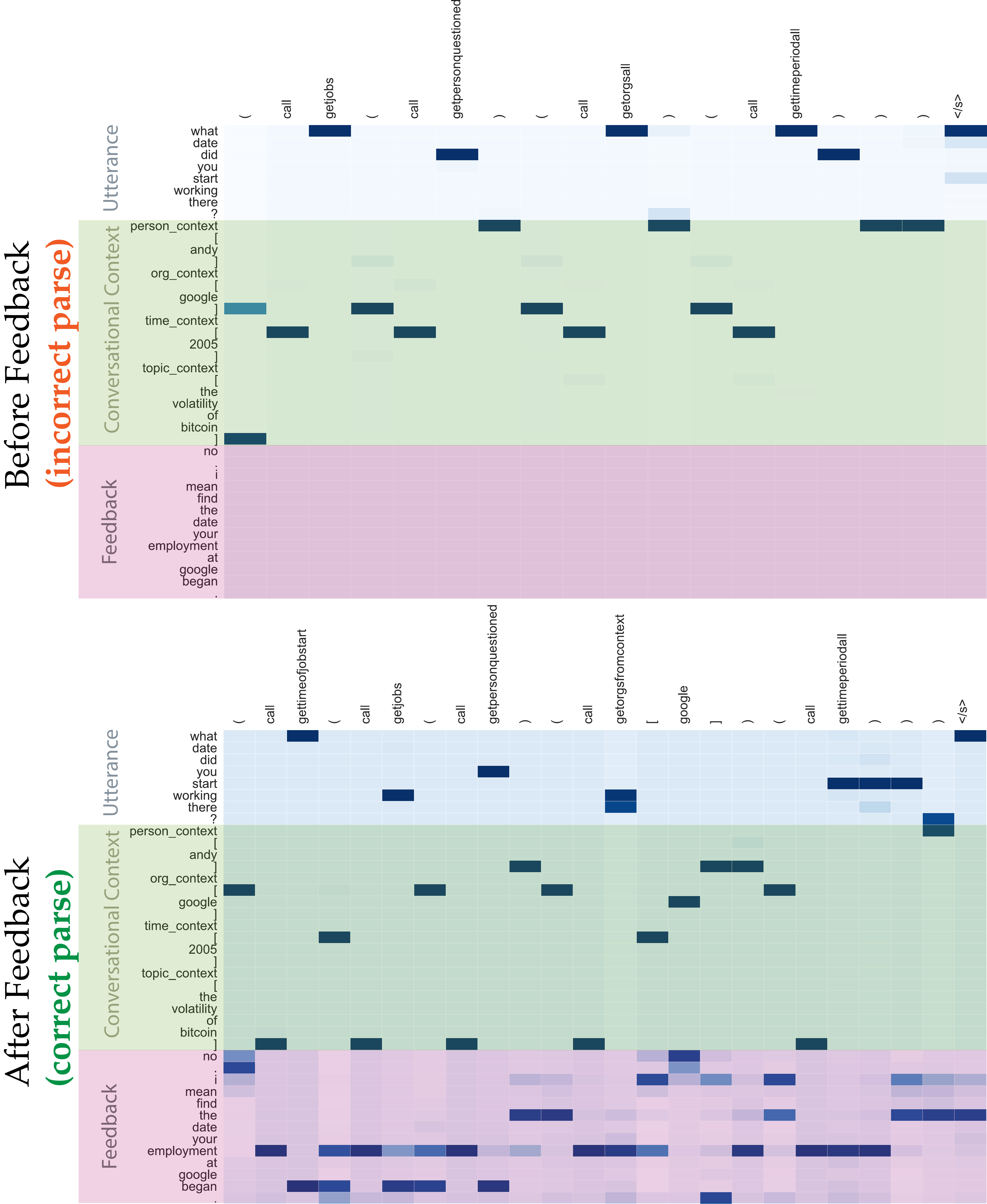}
\centering
\caption{Visualization of the attention mechanism in the task parser (top) and the feedback parser (bottom) for parsing the same utterance. We partition the input to the parser into three groups: the original utterance $u$ being parsed (blue), the conversational context (green) and the feedback utterance $f$ (red). This example was parsed incorrectly before incorporating feedback, but parsed correctly after its incorporation.}
\label{fig:combined_4}
\end{figure*}

\end{document}